\documentclass[letterpaper, 10 pt, conference]{ieeeconf} %
\IEEEoverridecommandlockouts               
\usepackage{flushend}
\usepackage[utf8]{inputenc}
\usepackage[T1]{fontenc}
\usepackage{bm}
\usepackage[super]{nth}
\usepackage{cite}
\usepackage{graphicx} 
\usepackage{subfigure}
\usepackage{textcomp,gensymb,amsmath}
\usepackage{adjustbox}
\usepackage{array,multirow}
\graphicspath{ {./Figure/} }
\usepackage{amsmath} 
\usepackage{amssymb} 
\usepackage{algorithm,algorithmic}
\title{\LARGE \bf
An Interaction-aware Evaluation Method for Highly Automated Vehicles
}

\author{Xinpeng Wang$^{1}$, Songan Zhang$^{1}$, Kuan-Hui Lee$^{2}$, Huei Peng$^{1}$ 
\thanks{$^{1}$Xinpeng Wang ({\tt\small xinpengw@umich.edu}), Songan Zhang, Huei Peng are with the Department of Mechanical Engineering, the University of Michigan, Ann Arbor, MI 48109, U.S.
}%
\thanks{$^{2}$Kuan-Hui Lee is with Toyota Research Institute, Los Altos, CA 94022, U.S.}%
}

\begin{document}

\maketitle
\thispagestyle{empty}
\pagestyle{empty}

\begin{abstract}
It is important to build a rigorous verification and validation (V\&V) process to evaluate the safety of highly automated vehicles (HAVs) before their wide deployment on public roads. In this paper, we propose an interaction-aware framework for HAV safety evaluation which is suitable for some highly-interactive driving scenarios including highway merging, roundabout entering, etc.
Contrary to existing approaches where the primary other vehicle (POV) takes predetermined maneuvers, we model the POV as a game-theoretic agent. To capture a wide variety of interactions between the POV and the vehicle under test (VUT), we characterize the interactive behavior using level-$k$ game theory and social value orientation and train a diverse set of POVs using reinforcement learning. Moreover, we propose an adaptive test case sampling scheme based on the Gaussian process regression technique to generate customized and diverse challenging cases. The highway merging is used as the example scenario. We found the proposed method is able to capture a wide range of POV behaviors and achieve better coverage of the failure modes of the VUT compared with other evaluation approaches.
\end{abstract}

\section{Introduction}
Highly automated vehicles (HAVs) are under rapid development all over the world. They have the potential to transform ground transportation by liberating people from tedious driving tasks and improving road safety by avoiding human errors. It's crucial to conduct verification and validation on their safety efficacy before their wide deployments. 

Safety evaluation of HAVs has been conducted at multiple traffic scenarios including unprotected left-turn, cut-in, and pedestrian crossing scenarios\cite{Zhao2017AcceleratedTechniques,Wang2019CombiningCrossing, Wang2020BehavioralVehicles}, etc. They can be characterized as reactive tests, where the vehicle under test (VUT) will be challenged by the primary other vehicle (POV) "in a surprise". The test case is fully defined by the initial condition of the challenge. Due to the short duration of the challenge, the POV is typically not programmed to interact with the VUT, and a predetermined trajectory is assumed. 

For SAE level 3 and above automated vehicles\cite{AutomatedNHTSA}, their operational design domain (ODD) can include dynamic and complex scenarios, including highway merging, roundabout entering, turning at unsignalized intersections, etc. For these scenarios, the existing methods show their limitations. For example, in highway merging, the VUT attempts to merge from the ramp onto the main road when another vehicle is present, which serves as the POV. The merging ramp gives enough time for the two vehicles to interact, and thus the assumption of "no interaction between POV and VUT" becomes unrealistic. Moreover, in the role of the POV, different human drivers may exhibit different behaviors under the same initial conditions, including coasting, accelerating to pull ahead, decelerating to yield, etc. These diverse behaviors pose a novel challenge for the motion prediction and decision-making module of the VUT, and should be incorporated into the evaluation framework.
 
We propose an interaction-aware evaluation methodology in this paper. It consists of two parts: first, we create a test case pool, in which we model a set of interactive POVs using level-$k$ game theory and social value orientation (SVO). Second, we propose an adaptive sampling scheme based on Gaussian process regression to generate challenging test cases for a given VUT. In this paper, we focus on the highway merging scenario, but this methodology can be applied to other scenarios including roundabout entering,  turning at unsignalized intersections, etc. 
This is the first effort to comprehensively identify the failure modes of a VUT in an interactive scenario.

The paper is organized as follows: Section 2 introduces related works; Sections 3 to 5 introduce the proposed method by first formulating the evaluation problem, then introduces the POV library construction and finally the adaptive test case generation procedure. Section 6 discusses the implementation details for the highway merging scenario; Section 7 shows the simulated testing results and the comparison with other sampling methods; finally concluding remarks are made in Section 8. 

\section{Related work}
Evaluation of the safety of HAVs has been an active area in recent years. Many test procedures have been proposed. Test matrix has been used to evaluate advanced driver assistance system (ADAS)\cite{NCAP2015EuropeanSystems}. However, the VUT can be tuned to pass the predefined test cases, but may fail under broader conditions in real-world driving tasks. Worst-case evaluation methods attempt to generate adversarial situations or POV inputs to create edge cases. \cite{Chou2018UsingDriving} and \cite{Althoff2018AutomaticVehicles} used reachability analysis to find test cases where the VUT has a minimal solution space. \cite{Tuncali2016UtilizingVehicles} applied simulation-based falsification to find failure cases for a given VUT. However, the assumption of adversarial POVs may not be reasonable. \cite{Zhang2020GeneratingVehicles} applied reinforcement learning to create adversarial yet socially acceptable POV behaviors in a highway driving scenario, while the diversity of the challenging scenarios has not been discussed. On the other hand, Monte-Carlo sampling-based evaluation methods have been proposed to generate test cases to estimate the real-world performance of the VUT. Some research uses importance sampling \cite{Zhao2017AcceleratedTechniques}\cite{Wang2019CombiningCrossing} or subset simulation \cite{Zhang2018AcceleratedTechnique} to efficiently estimate the crash rate of the VUT, while other works customize test cases to identify the failure modes of the VUT using adaptive sampling method \cite{Mullins2017AutomatedVehicles,Huang2017TowardsApproach, Feng2020TestingFramework}. The interactions between POV and VUT have not been considered in these works.

To model the interactive nature of human driving behavior, game theory has been widely applied, in which humans are modeled as utility-maximizing rational agents. Nash \cite{Schwarting2019SocialVehicles} or Stackelberg \cite{Fisac2019HierarchicalVehicles, Yoo2013AMerging} equilibrium models have been applied to model human driving behaviors. They rely on the assumption that each agent has an infinite level of rationality, which could be too strict considering that human drivers have to make quick decisions in a complex and dynamic environment. Therefore, other researchers assumed bounded rationality of human drivers and applied level-$k$ game theory \cite{Li2018GameSystems}, quantal response \cite{Sarkar2019AVehicles} or cumulative prospect theory \cite{Sun2019InterpretableTheory} to model human driving behaviors. On the other hand, \cite{Schwarting2019SocialVehicles} and \cite{Sun2018CourteousCars} considered the altruism of human driving behaviors in a game-theoretic setting. Despite the richness of game-theoretic models, they have yet to be comprehensively considered for HAV evaluations. Filling this gap is the focus of this work.

\section{Problem formulation}
\begin{figure}[]
  \centering
  \includegraphics[width=0.95\linewidth]{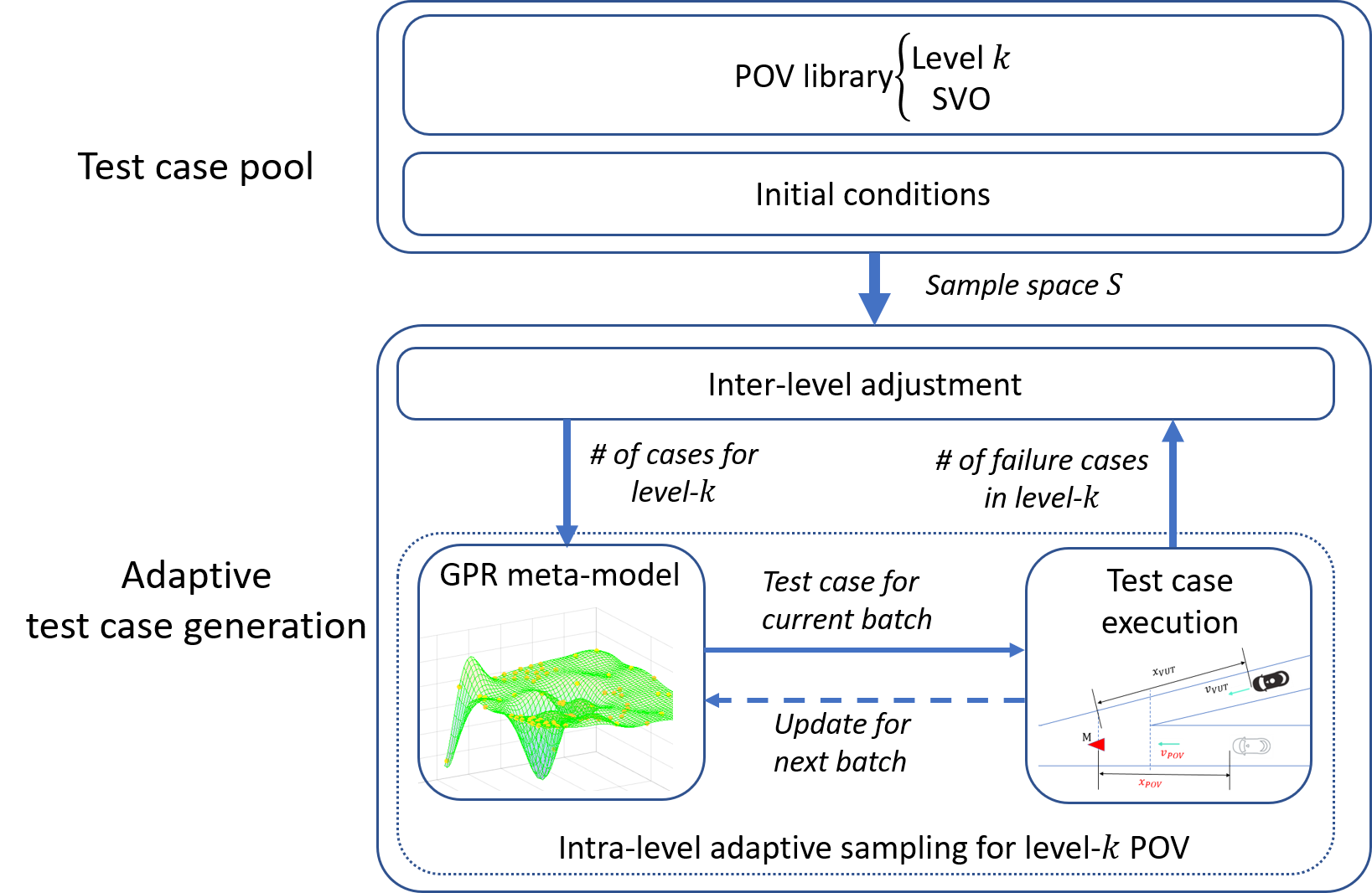}
  \caption{
  Pipeline of the interaction-aware evaluation method.}
  \label{fig:Outline}
\end{figure}

We aim to systematically generate test cases for a given VUT in interactive scenarios. The tasks are two-fold: firstly, we will create a test case pool for the target scenario; secondly, we proposed a mechanism to sample test cases from the test case pool. The test cases can be characterized by two sets of attributes: the first set defines the initial condition of the scenario; the second set describes the interactive and behavioral properties of the POV, which determines its driving policy. The test case sampling procedure aims to evaluate the safety performance of a black-box VUT by finding the failure modes of it through efficient sampling schemes. The overall concept of the proposed interaction-aware evaluation method is shown in Figure \ref{fig:Outline}.

\section{POV library construction}
The POV library needs to capture a diverse set of POV behaviors. On the one hand, the POV model should approximate the decision-making procedure of human drivers. Therefore, we assume that the POVs are game-theoretic agents, which take the (near) optimal action according to its utility function and assumptions of the opponents. On the other hand, the modeling framework should have the flexibility to describe a wide range of possible driving behaviors. We model the POVs as agents that possess different assumptions on the VUT and different utility functions for their own behavior. Specifically, we adopt the idea of level-$k$ game theory and social value orientation to describe the diversified POVs.

\subsection{Level-$k$ game formulation}
The level-$k$ game theory model is based on the idea that intelligent agents (such as human drivers) have finite level of rationality. The model first assumes a known level-0 agent, which is a naive agent that behaves non-cooperatively. Then, a level-$k$ agent ($k>0$) will assume that all the opponents are level-($k-1$) and will behave optimally according to this assumption. Using the level-0 policy as the starting point, the optimal policy for a level-$k$ agent can be generated sequentially. According to an experimental study in economics \cite{Costa-Gomes2009ComparingGames}, human decision-makers are usually as high as level-2 thinkers. Therefore, we only consider agents that are up to level-2 in this research to illustrate the concept. 
\begin{figure}[]
 \centering
 \subfigure[]{\label{fig:SVO} 
 \includegraphics[width=0.43\linewidth]{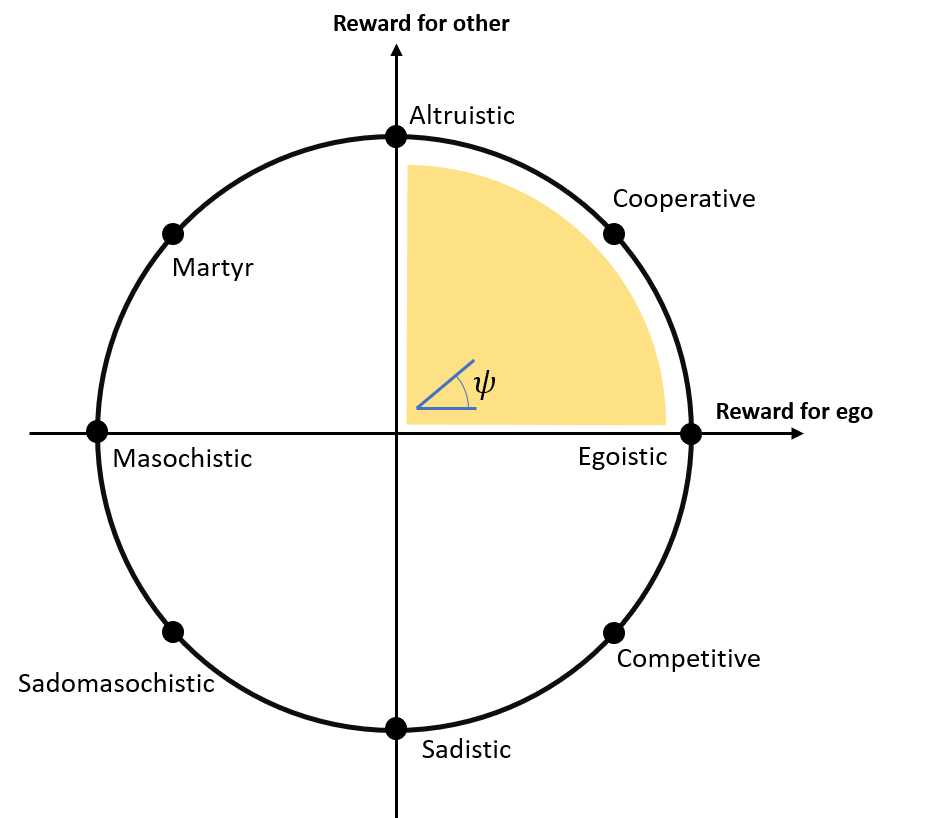}}
  \subfigure[]{\label{fig:levels_doubleHelix}
 \includegraphics[width=0.43\linewidth]{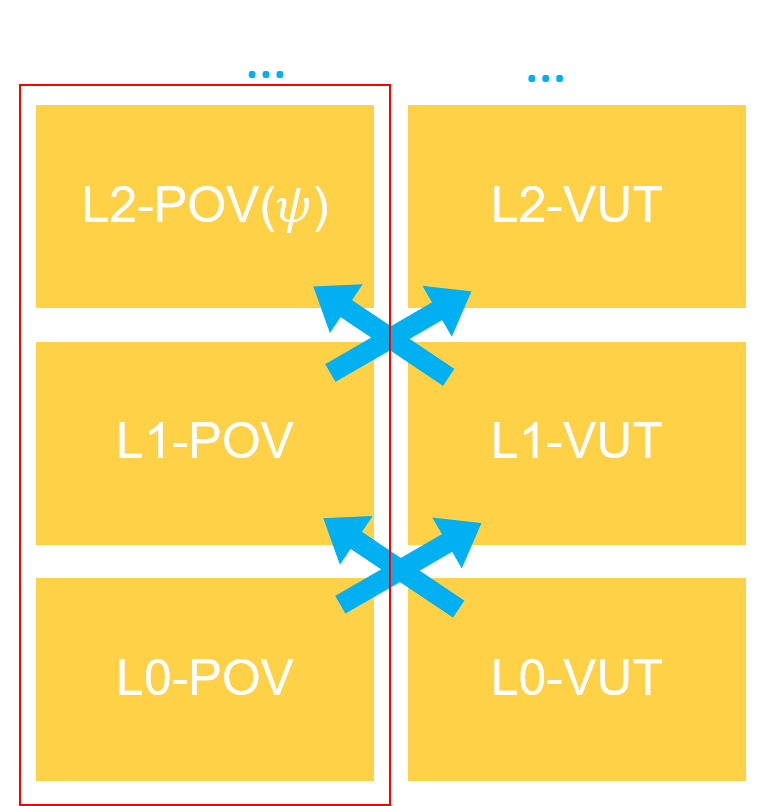}}
  \caption{(a) The SVO ring: we will focus on $\psi\in [0,\pi/2)$. (b) The hierarchy of level-$k$ agents. The level-2 POV has an extra parameter $\psi$ characterizing its SVO angle.}
  \label{fig:Model_POV} 
\end{figure} 
\subsection{Social value orientation}
Social value orientation (SVO) is a concept from social psychology literature, which quantifies the agent's degree of selfishness \cite{McClintock1989SocialBehavior}.  
It can be represented as an orientation angle $\psi$ indicating the agent's preference on the outcome for itself versus for others, as shown in Figure \ref{fig:SVO}, where different $\psi$ represent personalities including egoistic, pro-social, altruistic, competitive, etc. In a common game-theoretic setting, an agent is egoistic and will solely optimize for its own utility function, i.e. $\psi = 0$. However, when the variable SVO is combined with a game-theoretic driver model, as shown in \cite{Schwarting2019SocialVehicles}, it can significantly improve the accuracy of trajectory prediction for human drivers, thus better explain human driving behaviors. Moreover, agents with different SVO could represent a continuous spectrum of human drivers, which complements the level-$k$ framework where humans only have discrete types. In this work, we combine the SVO with the level-$k$ game theory to capture richer POV behaviors.
\subsection{POV library construction using reinforcement learning}\label{sec:POV library}
Based on the level-$k$ game theory and SVO, we create a library with the following POV agents: level-0 POV, level-1 POV and level-2 POV with varying social value orientation. Due to the non-competitive nature of driving tasks, we only consider the SVO angle in the $1^{st}$ quadrant for simplicity, i.e. $0\leq \psi < \pi/2$.
The reasons that we do not consider SVO for lower-level POVs are that: a level-0 POV is non-cooperative, thus SVO cannot be defined; a level-1 POV assumes its opponent is level-0 and non-cooperative, thus SVO is not defined either.

To construct the POV library, we first design the policy for a level-0 POV as a baseline. Next, to generate the policy for a level-$k$ POV $(k>0)$, a level-($k-1$) VUT is needed in advance. Therefore, we start with a level-0 VUT policy, and then generate higher-level POV and VUT sequentially in a double-helix structure, as shown in Figure \ref{fig:levels_doubleHelix}. Although the targets are level-$k$ POVs, we still need to compute level-$k$ VUTs as the stepping stones to obtain higher-level POVs. In simulated and real tests, the VUTs we evaluate are not these model VUTs.

For level-0 POV and VUT, they behave non-cooperatively with fixed speed profiles, which capture the behavior of inattentive drivers. For level-$k$ POV and VUT ($k>0$), we use reinforcement learning (RL) to compute their driving policies. To train a level-$k$ POV, we model it as an agent operating in an environment of level-($k-1$) VUT. The same procedure applies to VUT. To incorporate the factor of SVO, we consider the SVO angle as an extra state of the model when the level-2 POV is trained to generate a continuum of level-2 POVs. 

\subsubsection{Reinforcement learning basics}
Computing a rational agent can be modelled as an Markov Decision Process (MDP) problem, which is defined by $\mathcal{M}= (\mathcal{X}, \mathcal{U}, \mathcal{P}, \mathcal{R}, \gamma)$, with the state space $\mathcal{X}\subseteq \mathbb{R}^n$, the action space $\mathcal{U} \subseteq \mathbb{R}^m$, the transition dynamics of the environment $\mathcal{P}: \mathcal{X} \times \mathcal{U} \rightarrow \mathcal{X}$, the reward function $\mathcal{R} : \mathcal{X} \times \mathcal{U} \rightarrow \mathbb{R}$, and the discount factor $\gamma \in [0, 1)$. 

At each state $x_i$, an agent tries to compute a best action $u_i$ from the state-action mapping, i.e. the policy $\pi(x_i) = u_i$, that maximizes the expected cumulative reward, written as $E_{\pi}[ \sum\limits_{t=0}^{t_1} \gamma^t r(t)]$, where $t_1$ is the end time. To learn the optimal policy $\pi^*$, we use the Q-learning technique. We first define the action-value function $Q$:
\begin{equation}
  Q(x,u|\pi) = E_{\pi}\left[\sum\limits_{t=0}^{t_1}\gamma^t r_t|x_0=x,u_0=u \right]
\end{equation}
Then $\pi^*$ is learned by training the agent to learn the optimal $Q$ function, i.e. $Q^*(x,u|\pi^*)$, which satisfies the Bellman equation. For details please refer to \cite{sutton2018reinforcement}.

\subsubsection{Reinforcement learning formulation}
For a level-$k$ VUT, the state space includes all continuous physical states of the POV and the VUT, denoted as $X$ ($\mathcal{X} = X$). For POVs, the SVO angle is considered in the states space, which is held constant in each episode, i.e. $\mathcal{X} = X \times [0,\pi/2)$. The action space is a discrete set of acceleration or steering input. 

The reward function reflects the goal of driving for each agent. We assume that the reward function can be represented as:
\begin{equation}
  r(x,u) = W^T\Phi(x,u) = \sum\limits_{i=1}^k w_i\phi_i(x,u)
\end{equation}
which is a linear combination of multiple terms, each represents a different attribute for driving. There are three categories: 
\begin{enumerate}
  \item Ego reward for POV: $r_{POVe} = W^T_{POVe}\Phi_{POVe}$.
  \item Ego reward for VUT: $r_{VUTe} = W^T_{VUTe}\Phi_{VUTe}$.
  \item Safety reward for both: $r_{safe} = W_{safe}^T\Phi_{safe}$.
\end{enumerate}
The final reward function for VUT is: 
\begin{equation}
  r_{VUT} = r_{safe} + r_{VUTe}
\end{equation}
For a POV with SVO angle $\psi$, the reward function is:
\begin{equation}
   r_{POV} = r_{safe} + r_{POVe}\cos(\psi) + r_{VUTe}\sin(\psi)
\end{equation}
where $\psi$ modulates the rewards between POV and VUT. For a level-1 POV, $\psi \equiv 0$.
\subsubsection{Training POV \& VUT agents using DDQN}
In this work, since the state space is continuous, we use an artificial neural network as the function approximator for the optimal action-value function $Q^*$. The reinforcement learning algorithm we use is Double Deep-Q network (DDQN) \cite{VanHasselt2016DeepQ-Learning}. DDQN is based on the Deep-Q network (DQN) \cite{mnih2013playing} method. It addresses the problem of overestimating future return of DQN by decoupling the action evaluation and action selection into max operations in two different Q-networks. For the MDP with discrete action space and low dimensional state space, other advanced RL methods are not necessarily better than the DDQN approach. For other applications, DDQN can be replaced by other appropriate RL method.

\section{Adaptive test case generation}
\subsection{Problem formulation for adaptive testing}
From the previous section, we systematically generate the interactive POV library, which is characterized by the SVO $\psi$ and rationality level $L$. Combined with the initial condition of the scenario $x_0$, we can build the test cases pool, denoted as $\mathcal{S}$, where each case is $s=[x_0^T,\psi, L]^T$. Then, we need a mechanism to pick a set of $N$ test cases $\bm{s}=[s_1,...s_N]$ from the pool to identify the failure modes of the VUT. The main challenge is that different VUTs may have different performance profiles and weaknesses, and thus the failure modes are unknown at the beginning of the V\&V process. Therefore, we need a sampling scheme that can select new cases based on past test results to adaptively search for the weaknesses of each VUT as the testing proceeds. The goals of the test case generation process are two-fold: 
\begin{enumerate}
  \item Challenge: find cases where the VUT performs poorly (i.e. identify its weakness).
  \item Coverage: identify (possibly disassociated) regions of weak performance.
\end{enumerate} 

For a test run with case $s$, the performance of a VUT can be evaluated by function $P$, which takes the VUT trajectory $\bm{\tau}=[x(0),u(0),x(1),u(1)...x(t_1-1),u(t_1-1),x(t_1)]$ as input, and computes the performance score. It is written as:
\begin{multline}
  P(s) = f(\bm{\tau})=\mu_1I_{crash} + \mu_2 P_{safety}+\mu_3 P_{task} 
\end{multline}
where $I_{crash}$ is the indicator function for collision; $P_{safety}$ is the safety score; $P_{task}$ is the score on task accomplishment (success highway merge, smooth acceleration, etc); $\mu_1, \mu_2,\mu_3$ are weighting factors.

\begin{figure}[]
  \centering
  \includegraphics[width=0.8\linewidth]{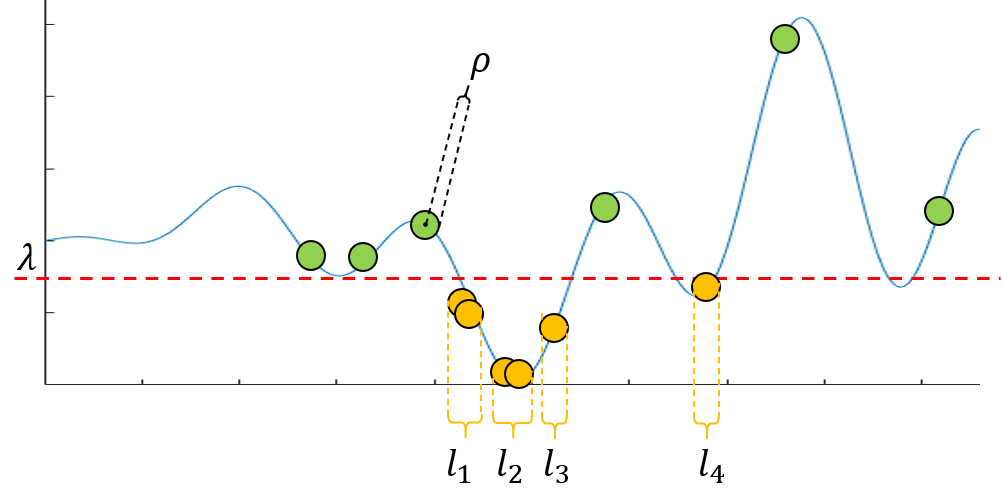}
  \caption{Measuring the FMC of test samples: the 1-D illustration. The blue curve represents the performance score $P(s)$, the red dashed line shows the performance threshold $\lambda$, and the regions of curve below the threshold are the failure modes. The FMC is computed as $M(\bm{s},\rho,\lambda) = l_1+l_2+l_3+l_4$.}
  \label{fig:Quality_sample}
\end{figure}

To describe the aforementioned two goals, we propose the criterion of failure mode coverage (FMC) for evaluating the quality of test samples:
\begin{equation}
  M(\bm{s},\rho,\lambda) = \int_{\bigcup B(\rho,\bm{s}_{I(\lambda)})} \bm{1} \,dv 
\end{equation}
where $B(\rho,s)$ is a hyper-ball centered around case $s$ with radius $\rho$; $\bm{s}_{I(\lambda)}$ is a subset of $\bm{s}$, such that $\forall s \in \bm{s}_{I(\lambda)}, P(s)<\lambda$.
The FMC evaluates the volume of the union of hyper-balls centered around test cases for which the VUT behaves poorly ($P(s)<\lambda$), which characterizes the coverage of failure modes for the VUT. Here, all the dimensions are normalized between [0,1]. Figure \ref{fig:Quality_sample} is a graphic illustration of the FMC in 1-D. 

\subsection{Adaptive testing method overview} 
To meet the goals of adaptive testing, we apply an adaptive sampling method. We generate $N$ test cases in batches sequentially, with batch size $n$. For each case in $\mathcal{S}$, the last attribute $L$ is a categorical variable, while all others are continuous variables, i.e. $\mathcal{S} = S \times \{0,1,2\}$. Therefore, we separate the sampling scheme for each batch into two stages, as shown in the lower part of Figure \ref{fig:Outline}. 
In the 1st stage, we allocate the number of samples into different POV levels, i.e. assign $n^{i}_{k}$ cases to be tested with level-$k$ POV at batch $i$. In the 2nd stage, we generate new test cases within each POV level from $S$ using Gaussian process regression (GPR). We will elaborate on the two stages later in this section. 

\subsection{Intra-level adaptive sampling}
Within each POV level, we conduct adaptive sampling using the Gaussian process regression (GPR). GPR is a non-parametric probabilistic model\cite{Rasmussen2006GaussianLearning}. The key idea is to maintain and update a GPR based meta-model based on existing samples, and use the meta-model to guide the generation of a new batch of samples. 
\subsubsection{Gaussian process regression}
Gaussian process (GP) is a stochastic process, for which the joint distribution of every finite collection of random variables follows a multivariate Gaussian distribution. A GP, as shown in (\ref{eqn:gp}), is characterized by its mean function $m(x)$ and a covariance function $k(x,x')$ (kernel).
\begin{equation}\label{eqn:gp}
  f(x) \sim GP(m(x),k(x,x'))
\end{equation}
In this work, we use GP to model the performance surface of each VUT, as shown in (\ref{eqn:gpr}).
\begin{multline}\label{eqn:gpr}
P(s)=\epsilon+f(s) \text{, where } \\ f(s)\sim GP(0,k(s,s'|\theta)),\epsilon \sim N(\beta,\sigma^2)  
\end{multline}
where $(\beta,\sigma,\theta)$ are the parameters of the model. In this work, we use a zero mean function and a square-exponential kernel function for the GPR model. Model parameters are optimized using maximum likelihood estimation. 
The procedure of adaptive sampling is illustrated in Algorithm \ref{alg:gpr}, and some details are explained below.

\begin{algorithm}
 \caption{Intra-level adaptive sampling}
 \label{alg:gpr}
 \hspace*{\algorithmicindent} \textbf{Input}: batches number $i$; batch size $n^{i}_{k}$; previous GPR model $\hat{P}^{i-1}_k$; exploration factor $\epsilon_0$. \\
 \hspace*{\algorithmicindent} \textbf{Output}: test cases with level-$k$ POV $\bm{s}_{k}^{i}$ and test results $\bm{y}^i_k$; updated GPR model $\hat{P}^{i}_k$.
 \begin{algorithmic}[1]
 \IF{$i = 1$}
  \STATE Sample initial test batch $\bm{s}_{k}^{1}$ uniformly from $S$.
 \ELSE
  \STATE Uniformly sample $p$ queries $\check{\bm{s}}$ from $S$ ($p>>n^{i}_{k}$).
  \STATE $\epsilon = \epsilon_0\alpha^{i-1}$.
  \STATE Pick $(1-\epsilon)n^{i}_{k}$ queries from $\check{\bm{s}}$ according to $q_{exploit}(s)$ as $\bm{s}_{exploit}$.
  \STATE Pick $\epsilon^{i}_{k}$ queries according to $q_{explore}(s)$ as $\bm{s}_{explore}$.
  \STATE $\bm{s}_{k}^{i} = [\bm{s}_{exploit}, \bm{s}_{explore}]$.
 \ENDIF
 \STATE Execute test cases $\bm{s}_{k}^{i}$, acquire results $\bm{y}^i_k$.
 \STATE Fit/Update the GPR model: $y = \hat{P}_k^{i}(s) = \hat{P}(s|\bm{s}^{1:i}_k,\bm{y}^{1:i}_k)$.
 \end{algorithmic}
\end{algorithm}

\subsubsection{Balancing between exploration and exploitation}
To achieve good coverage of the failure modes, we need to balance exploration and exploitation. On the one hand, it is desirable to explore regions with high uncertainty and pick more informative samples for more accurate meta-model, which helps on coverage. On the other hand, samples with low predicted $\hat{P}(s)$ represents more challenging cases, which are preferred for the goal of challenge. We attempt to solve this dilemma in an $\epsilon$-greedy way: on lines 6,7 of the Algorithm \ref{alg:gpr}, we evaluate the queries with two sets of query quality metric according to the GPR model, $q_{exploit}(s)$ and $q_{explore}(s)$:
\begin{align}
  q_{exploit}(s) = \hat{\mu}^{z_1}(s)\hat{\sigma}^{z_2}(s)\\
  q_{explore}(s) = \hat{\mu}^{z_3}(s)\hat{\sigma}^{z_4}(s)
\end{align}
where $\hat{\mu}(s) = \mathbb{E}[\hat{P}(s)]$, $\hat{\sigma}(s) = \mathbb{V}ar[\hat{P}(s)]$. 
$q_{exploit}(s)$ and $q_{explore}(s)$ have different parameters: the former prefers exploitation ($z_1>z_2$), and the latter focuses on exploration ($z_3<z_4$). For each batch, We pick the cases which maximize one of these metrics. The portion of cases for exploration and exploitation are determined by the parameter $\epsilon$. It which will gradually decrease across batches at the rate of $\alpha$ ($\alpha \in (0.9,1)$), such that the procedure starts with more exploration, and bias towards exploitation as more data are collected and a better meta-model is built.

\subsection{Inter-level ratio adjustment}
In this section, we will consider all the POV levels together by distributing cases into each level based on the results from the previous batch. Targeting on maximizing the expected coverage of the failure region, the strategy is to invest more samples in better-performing POV levels, while we keep exploring the other options. Specifically, we implement the following softmax decision rule on batch allocation:
\begin{equation}
n^{i+1}_{k} =  \pi_k^{i+1}n = \frac{\exp(\xi U(i,k))}{\sum\limits_{j=0}^{2}\exp(\xi U(i,j))}n
\end{equation}
where $U(i,k)=\frac{\text{\# of cases with } P(s)<\text{threshold}}{n^i_{k}}$, $\sum\limits_{j=0}^2{\pi_j^{i}} = 1$.
For the first batch, we distribute cases equally to all POV levels. After that, the cases are distributed according to the ratio of challenging cases found within that level in the previous batch. The parameter $\xi$ controls how "greedy" the decision rule is.

\section{Implementation on highway merging scenario}
\subsection{Scenario model}
\begin{figure}
  \centering
  \includegraphics[width=0.7\linewidth]{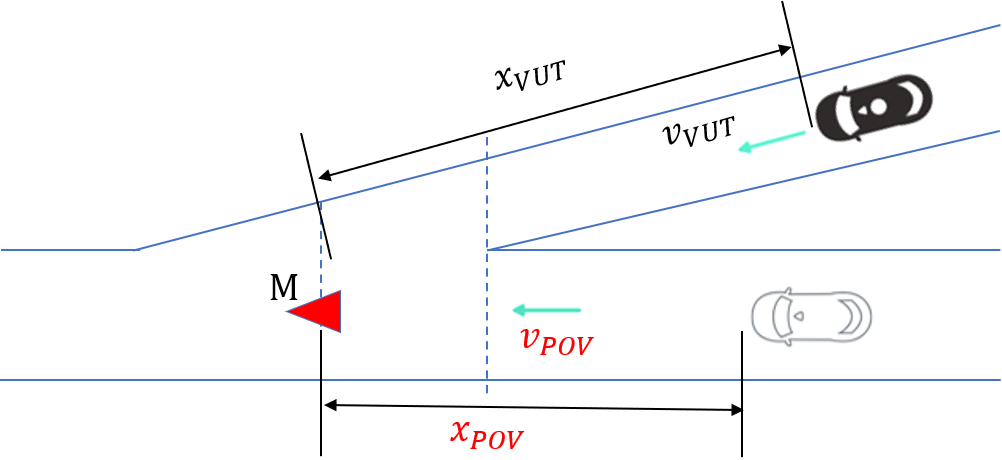}
  \caption{The configuration of highway merging scenario.}
  \label{fig:Merge_scene}
\end{figure}
The highway merging scenario is the focus of this paper, for which the configuration is illustrated in Figure \ref{fig:Merge_scene}. The VUT attempts to merge onto the highway, while the POV is driving on the main lane of the road. We make the following assumptions for simplification:
\begin{enumerate}
\item The POV and the VUT do see and interact with each other through the simulation horizon.
  \item The POV is not able to change lanes to yield to the VUT; the VUT can only merge at the merge point $M$, which is the origin for the lane-fixed coordinates for both the ramp and the main lane.
  \item There is only one POV on the main lane and there is no vehicle in front of the VUT on the ramp. 
  \item The scenario ends when the VUT reaches point $M$.
\end{enumerate}

We model both vehicles as double integrators and they only move longitudinally in their own lane. The equations of motion are:
\begin{equation}\label{eqn:eom}
\begin{cases}
  x_{POV}(t+1) = x_{POV}(t) + v_{POV}(t)\delta t \\
  v_{POV}(t+1) = v_{POV}(t) + a_{POV}(t)\delta t \\
  x_{VUT}(t+1) = x_{VUT}(t) + v_{VUT}(t)\delta t \\
  v_{VUT}(t+1) = v_{VUT}(t) + a_{VUT}(t)\delta t
\end{cases}
\end{equation}
where $x_{POV}, x_{VUT}$ are the longitudinal position, and $v_{POV}, v_{VUT}$ are the longitudinal speed of POV and VUT in their lanes. The input for each vehicle is the longitudinal acceleration, which ranges between $[a_{min}, a_{max}]$. The initial condition is characterized by $(x_{POV}^0, v_{POV}^0, x_{VUT}^0, v_{VUT}^0)$. Without loss of generality, we assume $x_{VUT}^0$ is fixed. Moreover, $v_{VUT}^0$ is observed rather than determined by the test conductor. Therefore, the initial condition to sample from is $x_0 = [x_{POV}^0, v_{POV}^0]^T$.
\subsection{Level-0 policy}
For the highway merging scenario, a level-0 POV is assumed to keeps a constant speed, regardless of the VUT. A level-0 VUT will accelerate with constant acceleration ($1m/s^2$) until the assumed highway speed ($28m/s$). 
\subsection{Training RL agents at the highway merging scenario}
When applied to the highway merging scenario, the physical state space of the MDP is $X=\mathbb{R}^4$, where each state is $x=[x_{POV}, v_{POV}, x_{VUT}, v_{VUT}]$. The transition dynamics are illustrated in (\ref{eqn:eom}), with the opponent's action governed by the level-($k$-1) policy. The actions are discrete acceleration choices within $a_{min}$ and $a_{max}$ for both POV and VUT, i.e. $u=a_{POV}/a_{VUT}\in U=\{-4,-3,...,0,+1,+2\} (m/s^2)$. Each episode terminates when the VUT reaches the merge point $x_{VUT}(t_1) = 0$. 

The detailed definitions of the three categories of reward mentioned in section \ref{sec:POV library} for the highway merging scenario are as follow:

$\Phi_{POVe}=[\phi_{acc}, \phi_{v_{HW}}]^T$, where $\phi_{acc}$ penalizes acceleration action; $\phi_{v_{HW}}$ penalizes speed exceeding the highway speed limits (either $v_{HWmin}$ or $v_{HWmax}$). The parameter values are shown in Table \ref{table:RewardInfo}.
 
 $\Phi_{VUTe}=[\phi_{acc}, \phi_{v_{min}},\phi_{v_{end}}]^T$, where $\phi_{acc}$ is the same as in $\Phi_{POVe}$; $\phi_{v_{min}}$ penalizes speed lower than a minimum speed $v_{min}$ during the episode; $\phi_{v_{end}}$ penalizes final merging speed of the VUT that is faster or slower than highway speed limits.
 
 $\Phi_{safe}=[\phi_{TTC}, \phi_{\Delta x}, \phi_{crash}]^T$ are the safety terms evaluated at the end of the episode $t_1$. We define: 
 \begin{align*}
   \Delta x_1 &= x_{POV}(t_1)-x_{VUT}(t_1) \\
   \Delta v_1 &= v_{POV}(t_1)-v_{VUT}(t_1) \\
   TTC &= \begin{cases}
     \frac{\Delta x_1}{-\Delta v_1} & when \; \Delta x_1\Delta v_1 < 0 \\
     \infty &otherwise
   \end{cases}
 \end{align*}
 where $\phi_{TTC}$ gives penalty when $TTC<TTC_{min}$; $\phi_{\Delta x}$ rewards large $|\Delta x|$, and gives penalty when $|\Delta x|<\Delta x_{critical}$; $ \phi_{crash}$ gives heavy penalty when $|\Delta x|<\Delta x_{crash}$.

 Finally, the DDQN algorithm for training the level-$k$ POVs and VUTs is implemented using the MATLAB reinforcement learning toolbox and Simulink.
\begin{table}
  \caption{Parameters for reward design}\label{table:RewardInfo}
  \centering
  \begin{tabular}{ | c | c | c | c |} 
  \hline
  $v_{HWmax}$ & 35.0 m/s & $v_{HWmin}$ & 24.6 m/s \\ 
  \hline
  $v_{min}$ & 12.0 m/s & $TTC_{min}$ & 7.0 s \\
  \hline
  $\Delta x_{crash}$ & 6 m & $\Delta x_{critical}$ & 15 m \\ 
  \hline
  \end{tabular}
\end{table}

\section{Simulation results}
We conduct interactive-aware testing to several baseline VUTs in simulations to validate the performance and benefits of the proposed method. 
\subsection{Baseline VUT}
For the highway-merge scenario, we design a rule-based algorithm for the merging vehicle (the VUT). Its decision-making has 3 stages:
\begin{enumerate}
  \item The VUT starts by following the speed profile of a level-0 VUT policy $\pi^0_{VUT}$. Go to stage 2 when it is $x^{rb}_1$ close to the merge point $M$.
  \item The VUT predicts $\Delta x$ relative to the POV when arriving at $M$, assuming the POV keeps a constant speed, and VUT follows $\pi^0_{VUT}$. If too close, switch to coast; else, keep following $\pi^0_{VUT}$. Go to stage 3 when VUT is $x^{rb}_2$ close to $M$ ($x^{rb}_2$ < $x^{rb}_1$).
  \item The VUT predicts $\Delta x$ with the POV when arriving at $M$, assuming the POV maintains a constant speed, and VUT follows $\pi^0_{VUT}$. If too close, switch to PID-control on acceleration; if not, follows $\pi^0_{VUT}$.
\end{enumerate}
By adjusting the parameters, we can manipulate the VUT to have different failure modes.
\subsection{Various interactive test cases}
\begin{figure}[]
 \centering
 \subfigure[Level-1 POV \& Level-0 VUT]{\label{fig:traj_1}
 \includegraphics[width=0.62\linewidth]{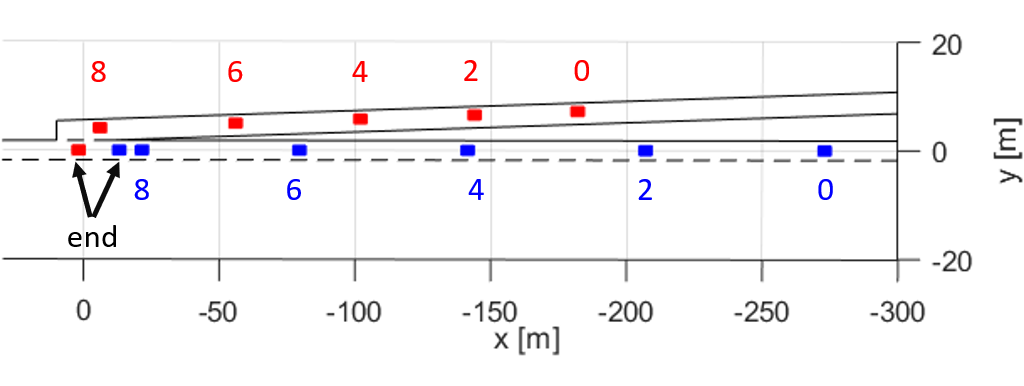}}
 \subfigure[]{\label{fig:traj_1_v}
 \includegraphics[width=0.33\linewidth]{  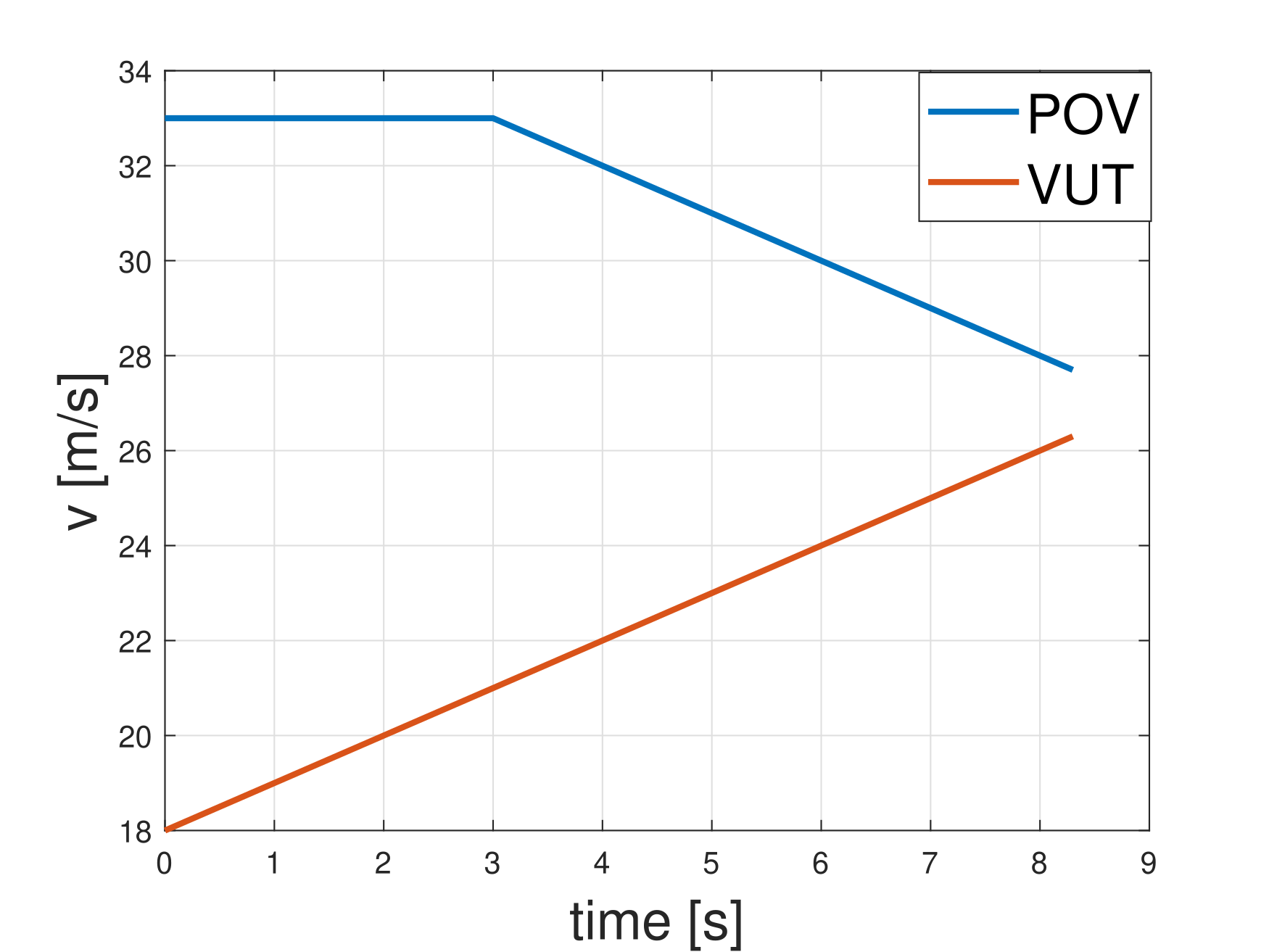}}
 \subfigure[Level-2 POV \& Level-1 VUT; $\psi = 0.60$]{\label{fig:traj_2} 
 \includegraphics[width=0.62\linewidth]{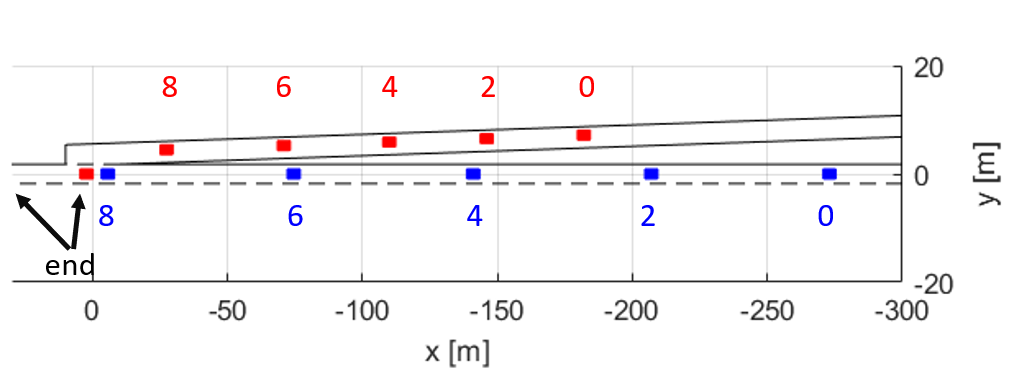}}
 \subfigure[]{\label{fig:traj_2_v}
 \includegraphics[width=0.33\linewidth]{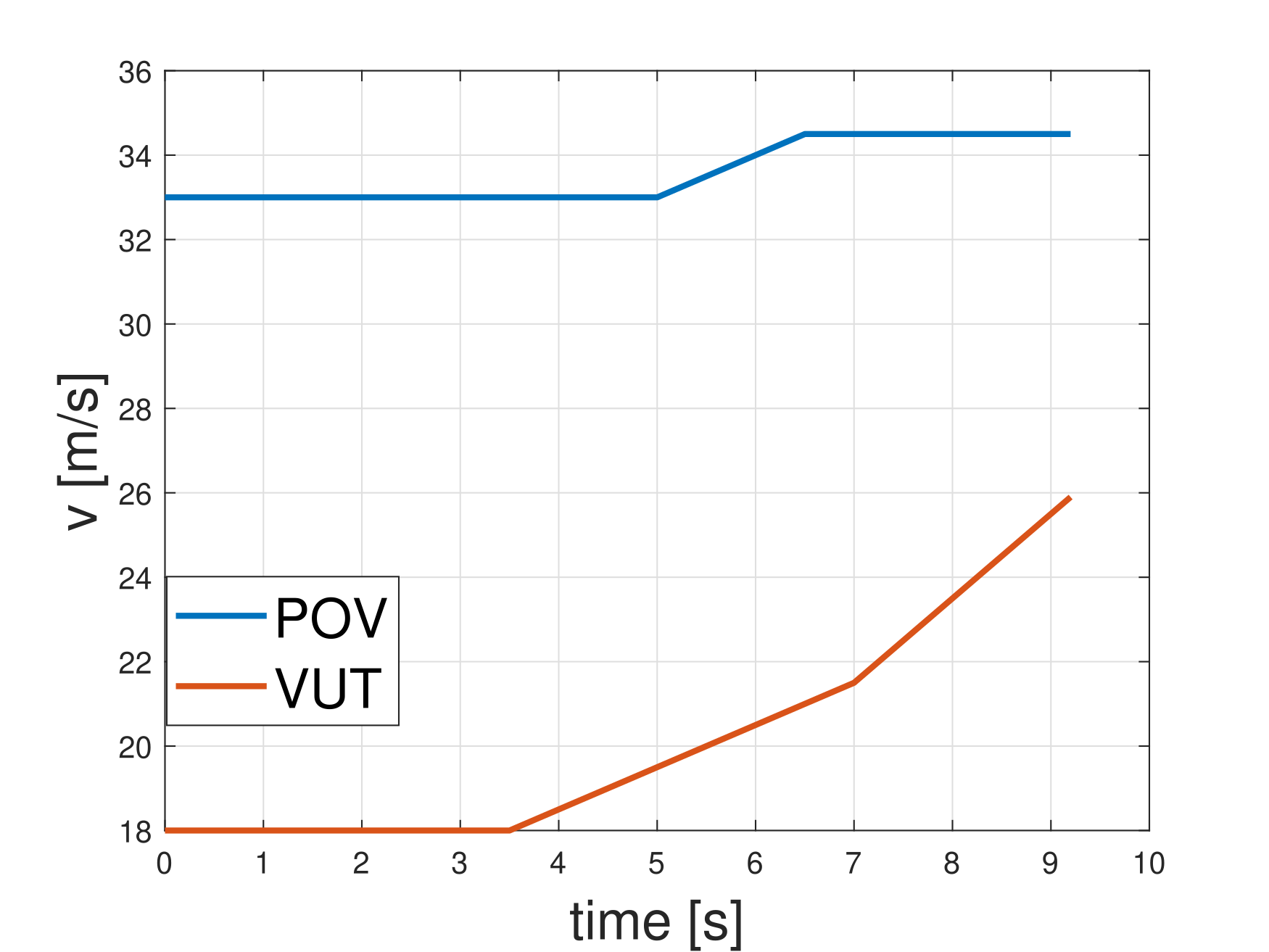}}
 \subfigure[Level-2 POV \& rule-based VUT \#2; $\psi = 0.60$]{\label{fig:traj_3} 
 \includegraphics[width=0.62\linewidth]{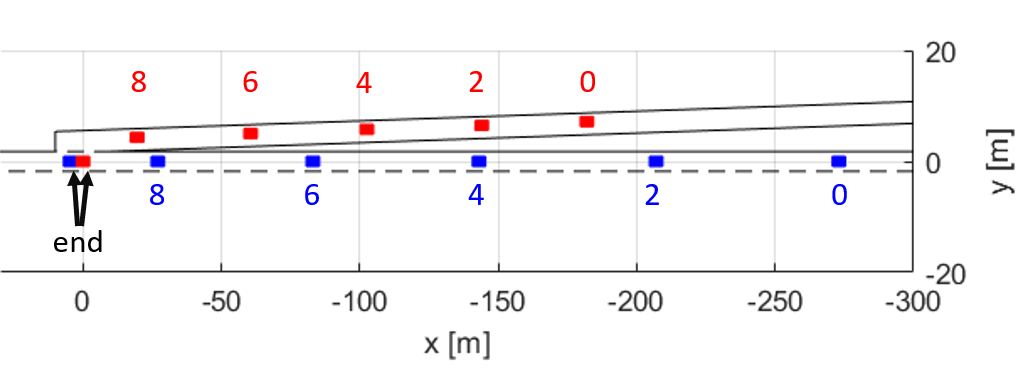}}
 \subfigure[]{\label{fig:traj_3_v}
 \includegraphics[width=0.33\linewidth]{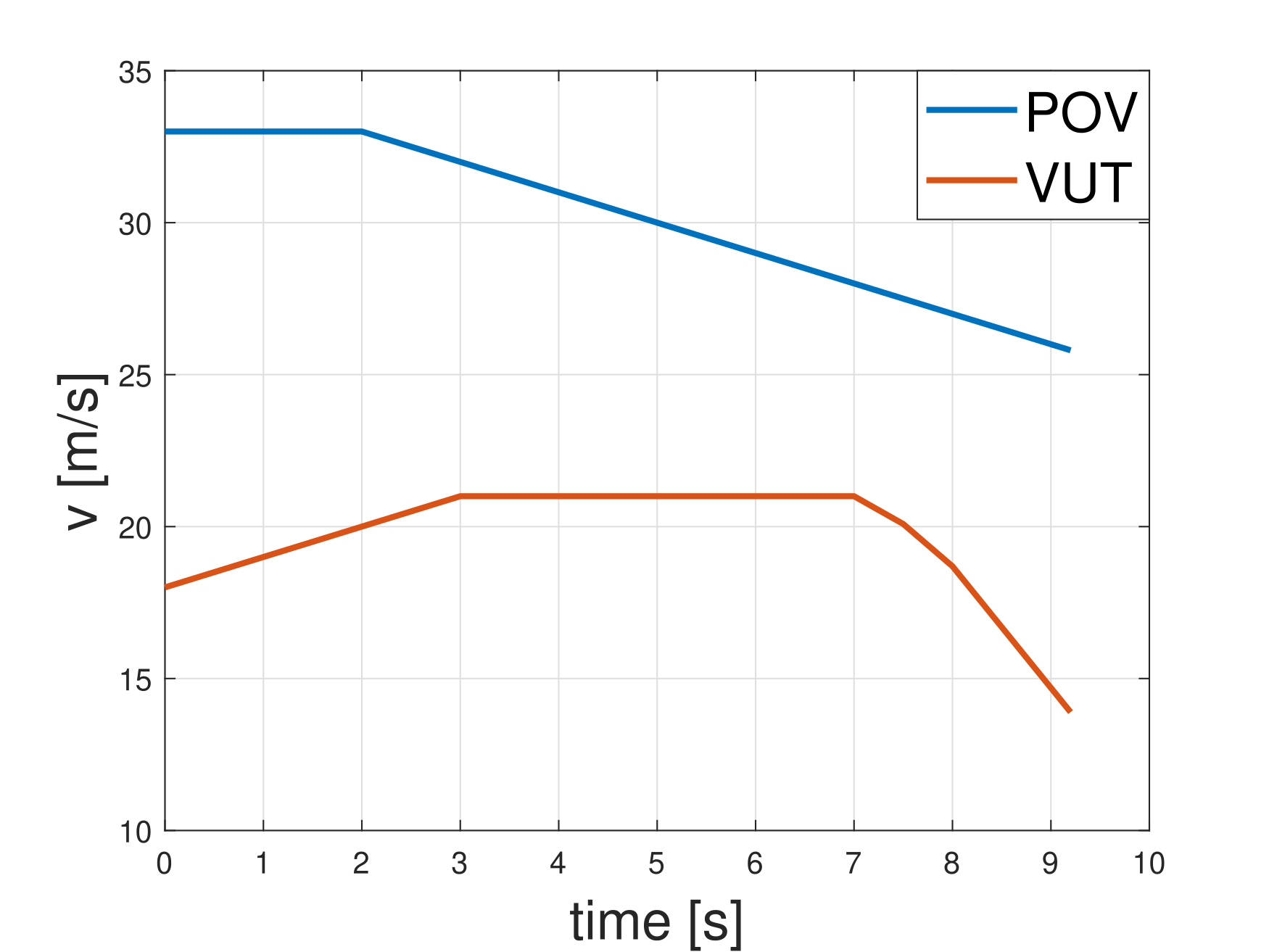}}
  \caption{Results with initial condition: $x_{POV}^0 = -273m$, $v_{POV}^0 = 33 m/s$, $v_{VUT}^0 = 18m/s$; blue for VUT, red for POV; the numbers show time lapses in seconds. }\label{fig:trajs_test_results}
\end{figure}

In this section, we present exemplar test cases with different interactions between the POV and the VUT. The road geometry of the highway merging scenario is based on an entrance ramp on US 23 North near exit 41. The VUT started at $x^0_{VUT} = -182 [m]$. In Figure \ref{fig:trajs_test_results}, we present three test cases with different POVs and VUTs. In all cases, the initial conditions are the same. In the 1st case, as shown in Figure \ref{fig:traj_1}, \ref{fig:traj_1_v}, the level-0 VUT is accelerating non-cooperatively, while the POV yields by reducing its speed to let the VUT merge first. In the 2nd scenario (Figure \ref{fig:traj_2}, \ref{fig:traj_2_v}), the level-1 VUT yields by starting its accelerating phase later, while the level-2 POV with a cooperative SVO accelerates to leave room for the VUT to merge behind. In the 3rd scenario, (Figure \ref{fig:traj_3}, \ref{fig:traj_3_v}), the same level-2 POV yields to let the VUT enter first. However, the rule-based VUT fails to understand the POV's intention. It starts to accelerate, then coasts and even decelerates hard before it crashes with the VUT. This last case shows a "stalemate" situation, when both agents try to yield to each other and create an inefficient and dangerous scene. These three test cases capture different interactions, which makes the evaluation scenarios diverse.

\subsection{Test results comparison}

\subsubsection{Results with a single POV level}
We first show the results of simulated testing with a fixed POV level. Each VUT is put through $N=400$ test cases. A case with a score $P(s)<-500$ means a collision has occurred, thus it is deemed as a failure case. We compare the proposed GPR-based adaptive sampling scheme to other test case generation schemes, including uniform sampling, simulated annealing \cite{dixon1978towards}, and subset simulation \cite{Zhang2018AcceleratedTechnique}. The FMC $M$ is the criterion for comparing their capability of discovering failure cases. For the VUT, two rule-based algorithm designs are selected, denoted as design \#1 and design \#2, where design \#2 has a faster response and is deemed "smarter". We test the two designs against Level-0 to Level-2 POVs. All methods are compared against the ground truth, which is generated by 10000 samples for level-0,1 and 20000 samples for level-2 POV using uniform sampling from the test case pool.
The quantitative results comparison is shown in table \ref{table:adaptiveResultComparision}. For all the three combinations of POV and VUT, the GPR-based adaptive sampling achieves the highest FMC among all the methods, and is also closest to the ground truth using only 4\% of the cases.  

\begin{figure}
 \centering
 \subfigure[Ground truth]{\label{fig:baseline_L1}
 \includegraphics[width=0.43\linewidth]{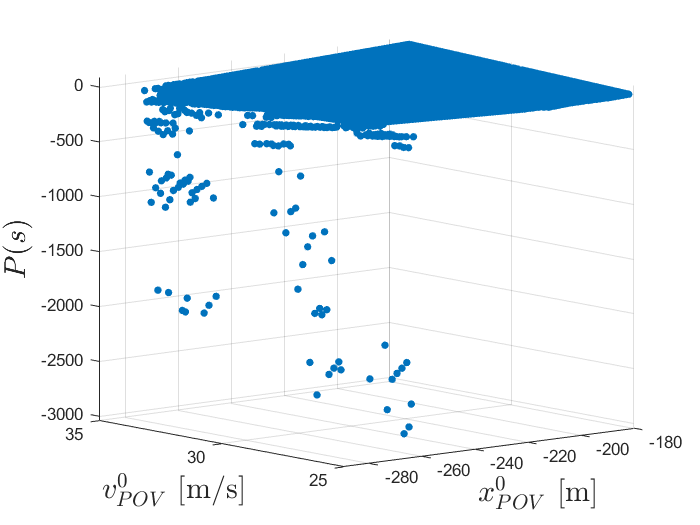}}
 \subfigure[GPR-based adaptive sampling]{\label{fig:GPR_L1} 
 \includegraphics[width=0.43\linewidth]{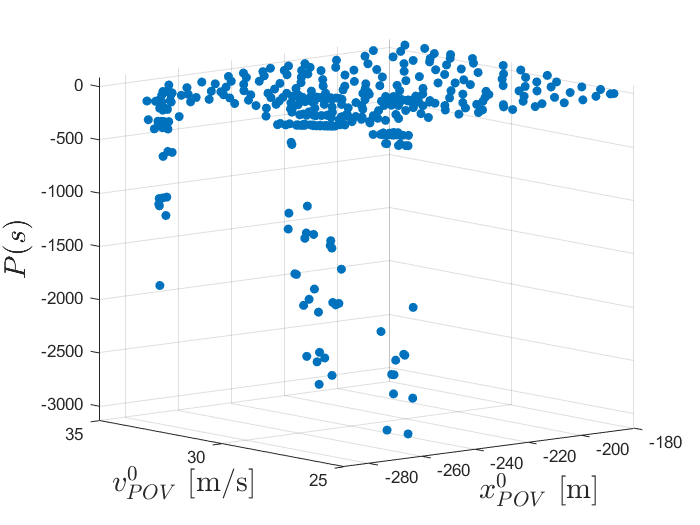}}
 \subfigure[Subset simulation]{\label{fig:SS_L1}
 \includegraphics[width=0.43\linewidth]{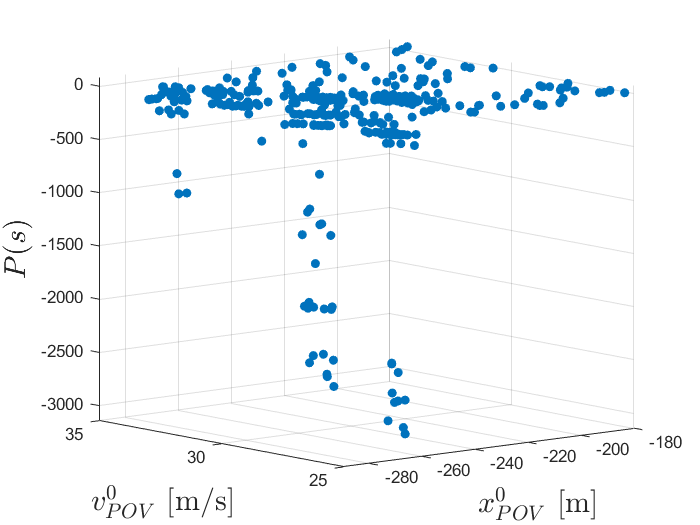}}
 \subfigure[Simulated annealing]{\label{fig:SA_L1}
 \includegraphics[width=0.43\linewidth]{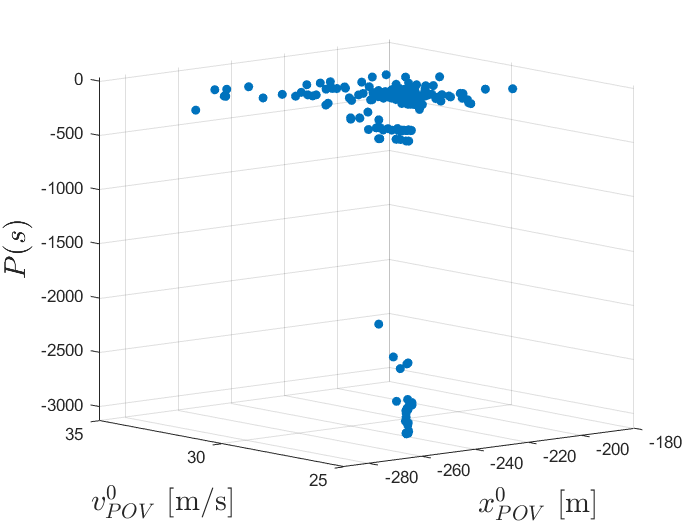}}
  \subfigure[Uniform sampling]{\label{fig:Uniform_L1}
 \includegraphics[width=0.43\linewidth]{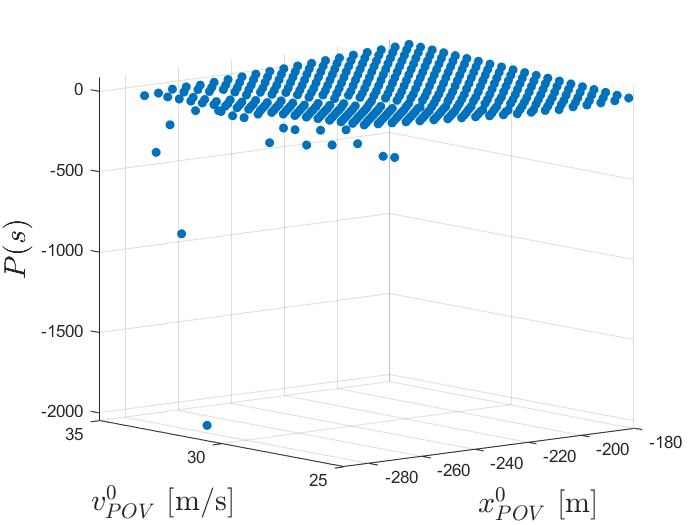}}
  \caption{Simulated testing results with level-1 POV on VUT design \#2; for (b)-(e), $k = 20$, $n = 20$.}\label{fig:L1_test results}
\end{figure}

\begin{figure}[]
 \centering
 \subfigure[Batch \#1]{\label{fig:gprCurve_1}
 \includegraphics[width=0.43\linewidth]{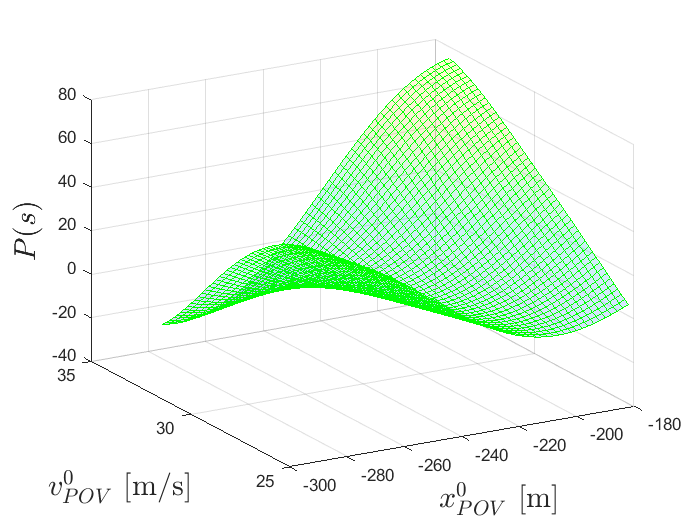}}
 \subfigure[Batch \#4]{\label{fig:gprCurve_4} 
 \includegraphics[width=0.43\linewidth]{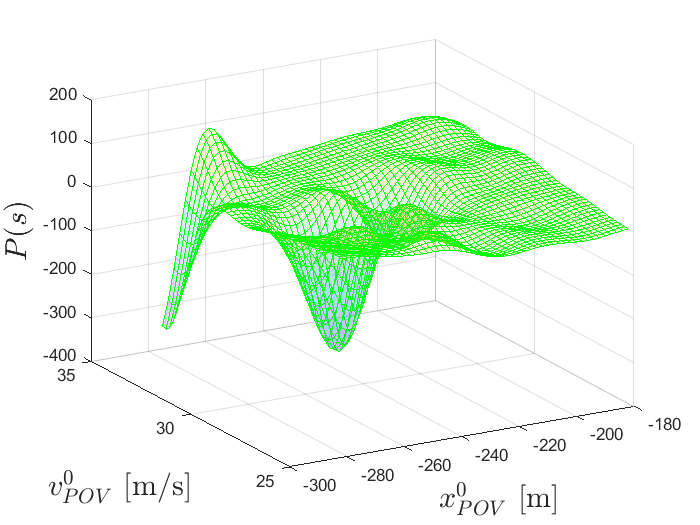}}
 \subfigure[Batch \#13]{\label{fig:gprCurve_13}
 \includegraphics[width=0.43\linewidth]{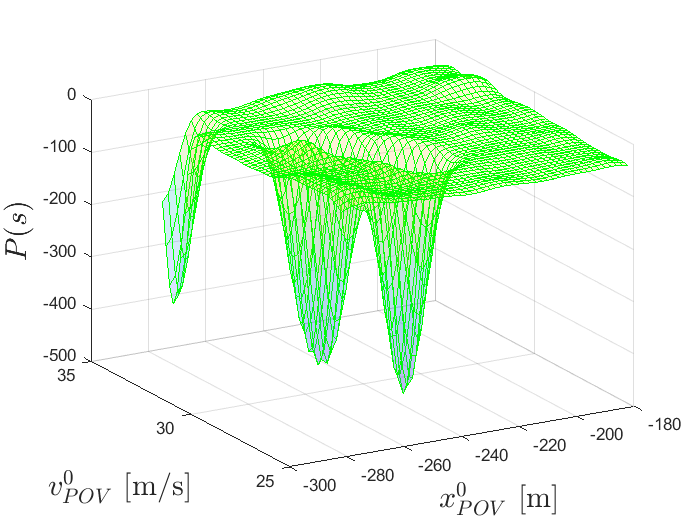}}
 \subfigure[Batch \#20]{\label{fig:gprCurve_20}
 \includegraphics[width=0.43\linewidth]{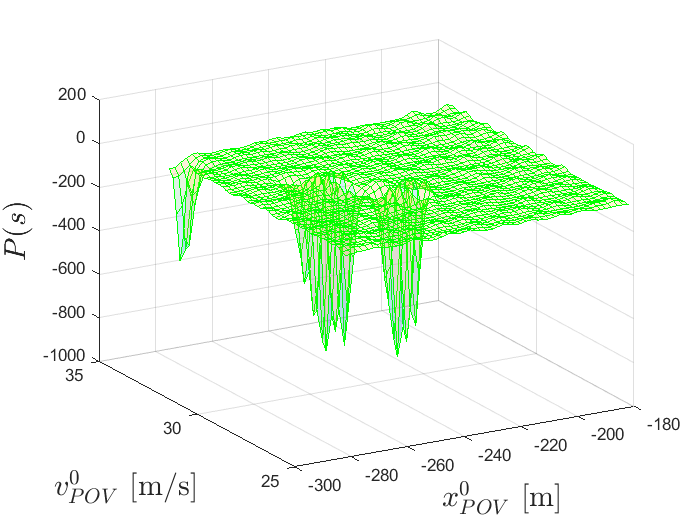}}
  \caption{The evolution of the GPR meta-model, with level-1 POV on VUT design \#2; $k = 20$, $n = 20$.}\label{fig:L1_curve}
\end{figure}

\begin{figure}[]
 \centering
 \subfigure[GPR-based adaptive sampling]{\label{fig:GPR_L2} 
 \includegraphics[width=0.43\linewidth]{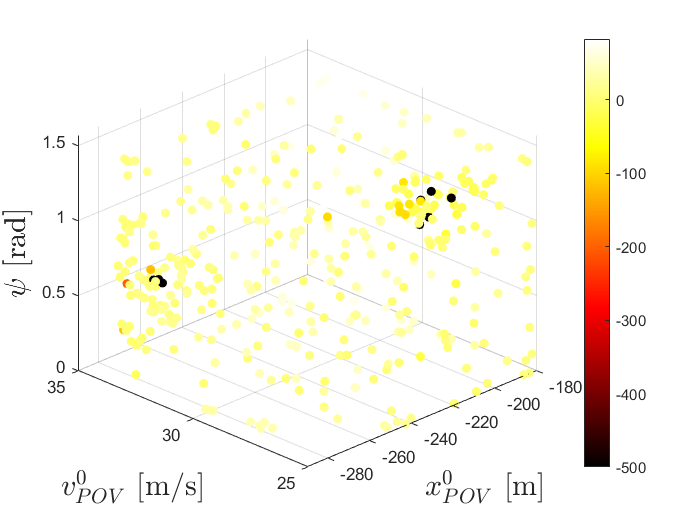}}
 \subfigure[Subset simulation]{\label{fig:SS_L2}
 \includegraphics[width=0.43\linewidth]{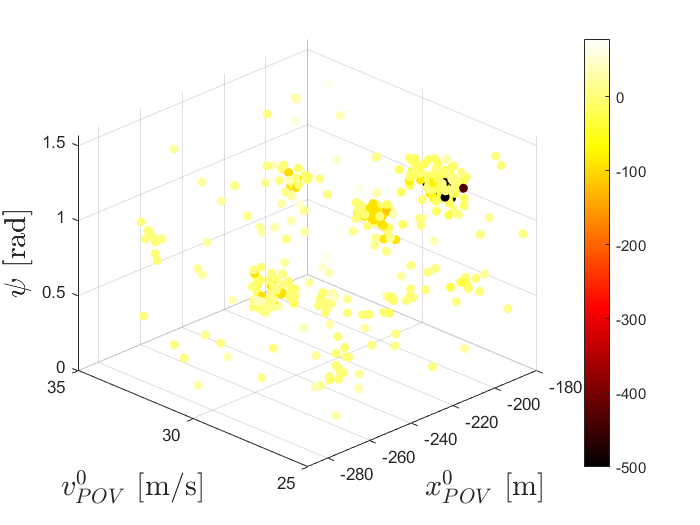}}
  \caption{Simulated testing results with level-2 POV on VUT design \#2; $k = 20$, $n = 20$.}\label{fig:L1_test result}\label{fig:L2_test results}
\end{figure}

Specifically, Figure \ref{fig:L1_test results} compares the results for testing VUT design \#2 against level-1 POV, which has three disjoint failure regions according to the ground truth. Uniform sampling can locate one failure region with very few failure cases; simulated annealing can find only one failure region, with many test cases concentrated around one local minimum; both subset simulation and GPR-based sampling can identify all three failure regions with only 4\% of the samples compared to the ground truth, while the GPR-based method achieves higher FMC and reconstructs the shape of the failure regions better. In Figure \ref{fig:L1_curve}, the progression of the GPR meta-model is displayed, where it finds more accurate failure modes as the batch number grows and test results accumulate. Figure \ref{fig:L2_test results} compares the results of testing with a level-2 POV. While the GPR-based method
can find the two failure modes far away from each other, subset simulation can only identify one of them with same number of tests. 
\begin{table}[]
  \caption{Adaptive testing results comparison}\label{table:adaptiveResultComparision}
  \centering
\begin{tabular}{ |c||c|c|c| }
 \hline
\multirow{3}{*}{Methods} & \multicolumn{3}{c|}{FMC $M(\bm{s},0.05,-500)$}\\ 
  \cline{2-4}
   & L-0 POV & L-1 POV & L-2 POV \\
   & \#1 VUT & \#2 VUT & \#2 VUT \\
 \hline
 Ground truth & 0.1296 & 0.0849 & 0.0073\\
 \hline
 Uniform sampling & 0.0223 & 0.0144 & 0\\
 \hline
 Simulated annealing & 0.0497 & 0.0184 & 0.0006\\
 \hline
 Subset simulation & 0.0830 & 0.0445 & 0.0019 \\
 \hline
 \textbf{GPR-based sampling} & 0.1012 & 0.0585 & 0.0032 \\
 \hline
\end{tabular}
\end{table}

\subsubsection{Results with multiple POV levels}
\begin{figure}[]
 \centering
 \subfigure[]{\label{fig:GPR_levels_allocation}
 \includegraphics[width=0.45\linewidth]{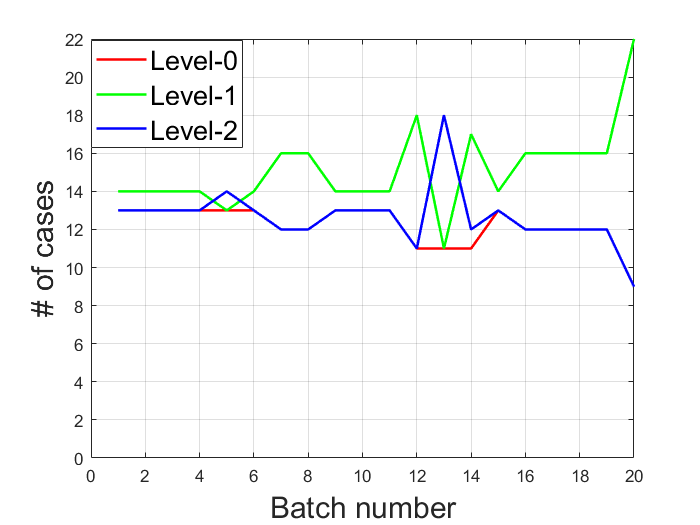}}
 \subfigure[]{\label{fig:GPR_levels_avgPs} 
 \includegraphics[width=0.45\linewidth]{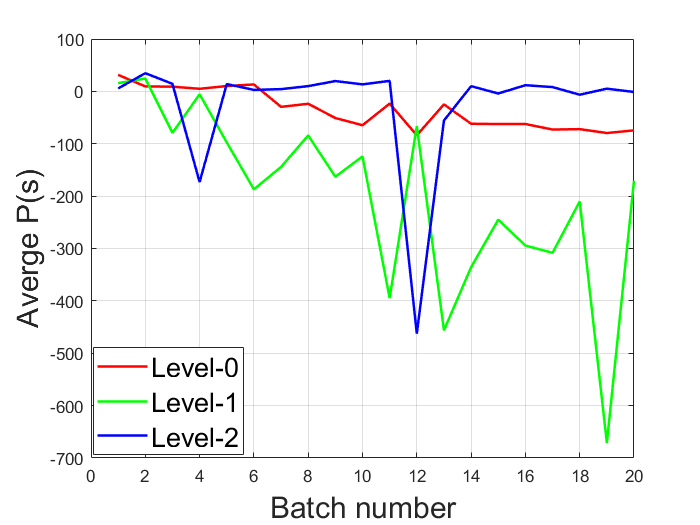}}
  \caption{Simulated testing results with the full POV library on VUT design \#2; $k = 20$, $n = 40$. (a) Batch sample allocation across POV levels in all batches. (b) Average performance score for each POV level in all batches. 
  }\label{fig:L012——results}
\end{figure}

Finally, we simulate the adaptive testing procedure with all the POV levels for VUT design \#2. The goal is to identify failure modes in all three levels within $N=800$ cases. Figure \ref{fig:L012——results} shows the change of sample allocation across different POVs. The sample sizes start evenly, but since more failure cases were found with level-1, and none for level-0 POV, The sample size grows for level-1 in later batches while shrinks for level-0. It is demonstrated that the proposed method is able to focus on the more promising interactive POV level for efficient identification of challenging test cases, while keeping exploring the under-performed ones. 

\section{CONCLUSIONS}
In this paper, we study the evaluation problem for black-box HAVs in scenarios with significant human interactions. We apply two game-theoretic methodologies, level-$k$ game theory and social value orientation, to model the interactive POV driving policies and incorporate them into our test case pool design. Then, we design an adaptive test case sampling scheme based on Gaussian process regression and propose a metric to assess failure mode coverage (FMC) to measure the test sample quality. We verify the proposed method by running simulated testing on several baseline VUTs. The POV library is able to emulate a wide variety of interactive behaviors and the sampling method can customize test cases to discover the failure modes of the VUTs by using only a fraction of the number of cases compared to the ground truth. It out-performs other sampling methods according to the FMC metric. 

\addtolength{\textheight}{-0cm}  

\section*{ACKNOWLEDGMENT}
Toyota Research Institute (TRI) provided funds to assist the authors with their research but this article solely reflects the opinions and conclusions of its authors and not TRI or any other Toyota entity.

We thank Shaobing Xu, Geunseob Oh, Yuanxin Zhong for their insightful suggestions and help.


\bibliographystyle{IEEEtran}
\end{document}